\documentclass[letterpaper, 10 pt, conference]{ieeeconf}  

\IEEEoverridecommandlockouts                             
\usepackage{graphics}
\usepackage{epsfig} 
\usepackage{mathptmx} 
\usepackage{times} 
\usepackage{amsmath} 
\usepackage{amssymb}  
\usepackage{graphicx}
\usepackage{float}
\usepackage{tabularray}
\usepackage{bm}
\usepackage[dvipsnames]{xcolor}
\usepackage[caption=false]{subfig}
\usepackage{mathtools}
\usepackage{stfloats}
\usepackage{mathrsfs}
\usepackage{multirow}
\usepackage{makecell}
\usepackage{vcell}
\usepackage{booktabs}
\usepackage{diagbox}
\usepackage{csquotes}
\usepackage{cite}
\usepackage{tikz}
\usepackage{lipsum}
\usepackage{schemata}

\usepackage[hyphens]{url}
\usepackage[hidelinks]{hyperref}
\hypersetup{breaklinks=true}
\usepackage{cite}

\usepackage{algorithm}
\usepackage[noend]{algpseudocode}

\algrenewcommand\algorithmicforall{\textbf{for each}}
\algrenewcommand\algorithmicindent{.8em}

\algdef{SE}[DOWHILE]{Do}{doWhile}{\algorithmicdo}[1]{\algorithmicwhile\ #1}%
\algnewcommand\algorithmicforeach{\textbf{for each}}
\algdef{S}[FOR]{ForEach}[1]{\algorithmicforeach\ #1\ \algorithmicdo}

\usepackage{hyperref}
\hypersetup{
colorlinks=true,
linkcolor=blue,
filecolor=magenta,
urlcolor=blue,
}

\title{\LARGE \bf
ZeroSCD: Zero-Shot Street Scene Change Detection
}
\author{Shyam Sundar Kannan and Byung-Cheol Min
\thanks{The authors are with SMART Lab, Department of Computer and Information Technology, Purdue University, West Lafayette, IN 47907, USA \tt{\small{kannan9@purdue.edu | minb@purdue.edu}}}%
}

\begin{document}
\maketitle
\thispagestyle{empty}
\pagestyle{empty}

\begin{abstract}
Scene Change Detection is a challenging task in computer vision and robotics that aims to identify differences between two images of the same scene captured at different times. Traditional change detection methods rely on training models that take these image pairs as input and estimate the changes, which requires large amounts of annotated data, a costly and time-consuming process. To overcome this, we propose ZeroSCD, a zero-shot scene change detection framework that eliminates the need for training. ZeroSCD leverages pre-existing models for place recognition and semantic segmentation, utilizing their features and outputs to perform change detection. In this framework, features extracted from the place recognition model are used to estimate correspondences and detect changes between the two images. These are then combined with segmentation results from the semantic segmentation model to precisely delineate the boundaries of the detected changes. Extensive experiments on benchmark datasets demonstrate that ZeroSCD outperforms several state-of-the-art methods in change detection accuracy, despite not being trained on any of the benchmark datasets, proving its effectiveness and adaptability across different scenarios.

\end{abstract}

\section{Introduction}
\label{sec:intro}
\begin{figure}[t]
    \centering
    \includegraphics[width=\linewidth]{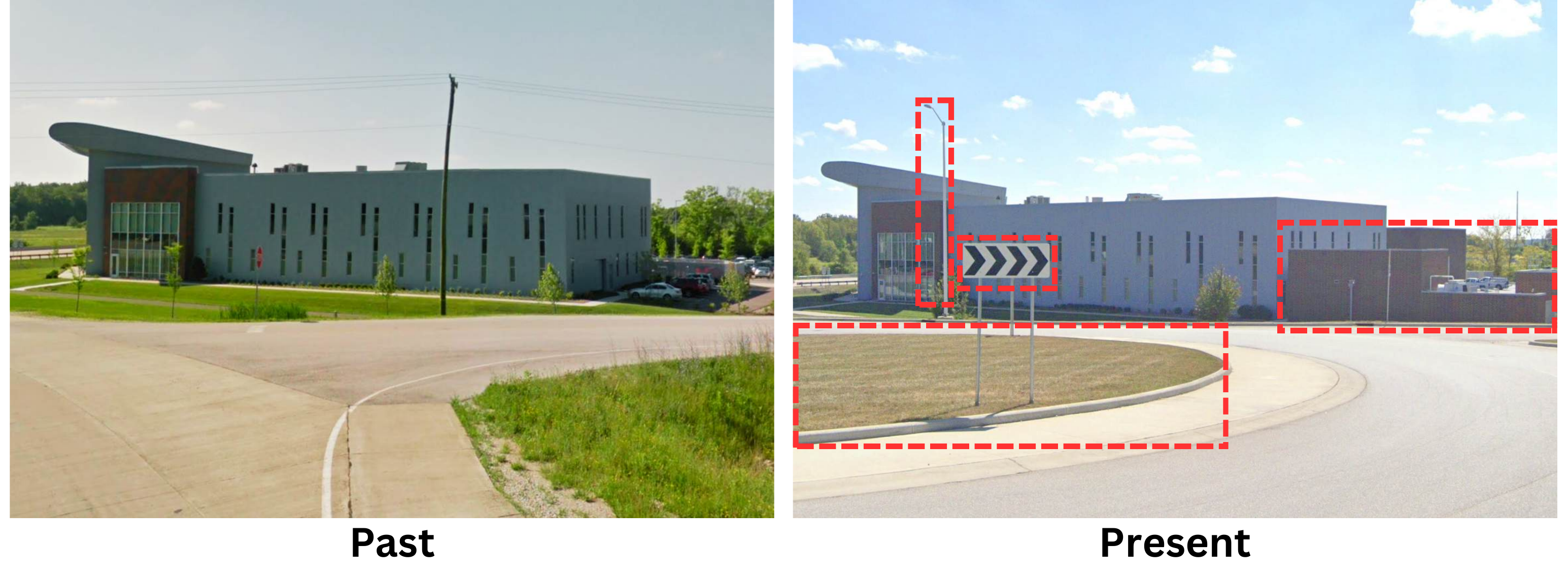}
    \vspace{-20pt}
    \caption{Two images of the same location, taken before and after the construction of a roundabout, are shown. The areas where changes have occurred are highlighted with a \textcolor{red}{red} box. Detecting these changes related to the roundabout and signs is essential to ensure the map is updated for the safe navigation of autonomous vehicles. }  
    \vspace{-10pt}
    \label{fig:concept_pic}
\end{figure}

In robotics and autonomous vehicles, the operational environment of an autonomous agent is frequently subject to geometric and structural changes due to both natural phenomena and human-made factors. These changes can arise from events such as natural disasters, like earthquakes, or the construction of new buildings. It is crucial for autonomous robots and vehicles to detect these changes and continuously update the environmental map of their operational space \cite{zhang2021real, boubakri2022high}. Failure to detect and update these changes can compromise the accuracy of localization and navigation, leading to potential safety and efficiency issues. Fig.~\ref{fig:concept_pic} shows two images of the same location captured at different times and a roundabout and additional buildings have been constructed recently. Detecting these changes and updating the map is essential for appropriate path planning in autonomous vehicles. Therefore, detecting such structural changes is essential for maintaining the safe and efficient operation of autonomous agents.

Street Scene Change Detection (SCD) is a significant problem in computer vision that focuses on identifying changes between two street view images of the same scene or location, captured at different times \cite{radke2005image}. These temporal gaps allow for various structural changes in the environment, which SCD aims to detect. Beyond structural changes, the images may also exhibit style variations such as changes in viewpoint, illumination, weather, and seasons \cite{lowry2015visual}. Therefore, SCD techniques must be capable of accurately detecting structural changes while remaining robust to these non-structural variations. While SCD plays a vital role in robotics, it also finds applications in traffic monitoring \cite{wan2020intelligent}, real estate assessment \cite{du2012fusion}, and disaster evaluation \cite{jst2015change}.

Deep learning has emerged as a widely used approach for addressing SCD. These methods typically involve training on annotated datasets consisting of image pairs: one captured before and the other after the change, along with a binary mask highlighting the changed regions. The success of these methods heavily depends on both the quality and availability of data. However, they face two main challenges: data scarcity and vulnerability to variations. First, creating an SCD dataset is particularly challenging, as it requires capturing images of the same location over time and manually annotating the changes, which is a labor-intensive process. To mitigate this, semi-supervised \cite{lee2024semi, sun2022semisanet} and self-supervised \cite{seo2023self, furukawa2020self} methods have been proposed, reducing the need for manual labeling. Nonetheless, the cost of data collection remains. Second, images captured over time can undergo significant style variations due to environmental factors such as weather and season. Therefore, SCD models must be robust to these style changes while detecting structural modifications. However, datasets often fail to capture all possible style variations, which limits the generalization of trained models to real-world scenarios where these variations are common.

To overcome these limitations, we propose ZeroSCD, a zero-shot, training-free framework for scene change detection. By eliminating the need for training, ZeroSCD reduces the time and cost associated with data collection and annotation. In SCD, Visual Place Recognition (VPR) is typically used to pair the current scene with a previously captured image of the same location. VPR facilitates image retrieval by identifying the closest match to the current scene, forming the foundation for change detection. VPR models are generally trained to extract robust, style-invariant features that can withstand variations in lighting, weather, and season \cite{masone2021survey}.

ZeroSCD leverages our previous VPR model, PlaceFormer \cite{kannan2024placeformer}, which is specifically designed to extract features resilient to these style changes. Building on this robust foundation, ZeroSCD efficiently detects structural changes while remaining resistant to non-structural variations. The core idea of ZeroSCD is to utilize the robust features extracted by the VPR model to establish correspondences between the images. These are then combined with the output from a foundational semantic segmentation model to precisely identify and localize the changes. More specifically, VPR-derived features are used to detect changes between images, while the segmentation model helps define the precise boundaries of the altered objects. This combination ensures that ZeroSCD can accurately detect and localize changes, even in the presence of style variations, without requiring any additional training or annotated data.

In summary, the main contributions of our work are:

\begin{itemize} 
    \item A novel zero-shot change detection framework, ZeroSCD, capable of detecting changes between images captured at different times without requiring any training.
    \item The method is built upon a robust Visual Place Recognition model, which ensures resilience to style variations such as changes in lighting, weather, and season, enabling more accurate detection of structural changes. 
    \item Extensive validation of ZeroSCD across multiple change detection datasets, where it achieves state-of-the-art performance on several benchmarks, despite not being trained on any of them. 
\end{itemize}


\section{Related Works}
\label{sec:rel_work}
\subsection{Change Detection}
Change detection involves the comparison of two images captured at different times to identify alterations or transformations that have occurred over a given period. Traditionally, this task was performed at the pixel level by extracting features for each individual pixel \cite{ilsever2012pixel}. However, pixel-based methods rely heavily on precise alignment (registration) of the image pairs, making them highly susceptible to errors caused by misregistration. Another traditional approach is object-based change detection methods where the changes are detected at an object level rather than at pixel level \cite{huo2009fast, im2008object}. These approaches typically employ segmentation algorithms to detect objects within the images and then track changes across the detected objects. However, these methods are not easily scalable and often struggle under varying conditions such as changes in illumination, weather, and viewpoint, limiting their applicability in dynamic real-world environments.

Deep learning has substantially advanced change detection by improving feature extraction processes, leading to more accurate identification of changes between images. Unlike traditional methods, deep learning models produce features that are inherently robust to style variations. The development and training of these models have been greatly facilitated by specialized change detection datasets, including VL-CMU-CD \cite{alcantarilla2018street}, Tsunami, and Google Street View (GSV) \cite{jst2015change}. Several feature embedding-based methods \cite{varghese2018changenet, zhan2018log, guo2023local} have been proposed to address change detection tasks. These approaches typically employ an encoder-decoder architecture, where images are first encoded to extract features and then decoded using a detection head to produce a binary change mask that highlights differences between the images. While effective, these methods assume a perfect one-to-one match between the images, and their performance can degrade when such an exact match is not present. To address this, methods incorporating a warping module to align the images more accurately have also been introduced \cite{park2022dual}. However, these approaches are often data-intensive, requiring extensive annotated datasets for training, and may lack generalizability, necessitating retraining for different scenes or localities, which limits their scalability.

To address issues related to data availability, semi-supervised approaches have emerged as a promising alternative, allowing models to be trained with minimal or no labeled data \cite{lee2024semi, sun2022semisanet, seo2023self}. While these methods reduce the dependency on fully annotated datasets, they still require substantial amounts of data, which can be challenging to collect over extended periods. Moreover, these approaches often struggle to generalize across diverse scenes or terrains, limiting their effectiveness in varied environments.

To overcome these limitations, zero-shot change detection methods have been proposed, which aim to detect changes between images without any prior training \cite{cho2024zero}. For example, \cite{cho2024zero} frames the change detection problem as a tracking task, employing a pre-trained tracking model to identify changes. While this represents a significant step towards zero-shot change detection, the performance of such methods can be inferior to supervised approaches. This is because tracking models are typically designed for tracking objects across consecutive video frames, rather than detecting significant displacements or changes in static images, which poses challenges for scenarios requiring large spatial transformations. Therefore, in this work, we propose a zero-shot change detection framework that leverages features extracted from a VPR model. By utilizing VPR features, our approach benefits from inherent robustness to style variations, and these features facilitate accurate estimation of correspondences between the two images, which is crucial for effective change detection. 

\subsection{Segment Anything Model}
Segment Anything Model (SAM) \cite{kirillov2023segment} is a vision transformer-based framework designed to handle diverse image segmentation tasks with remarkable versatility. SAM can generate segmentation masks for virtually any object, including those it has not encountered during training, making it highly adaptable across various domains. Given an image, SAM produces hierarchical segmentation masks at different levels of granularity, enabling precise object segmentation. In this work, we use the segmentation masks from SAM to extract precise boundaries of various objects and couple them with the changes detected using VPR features for accurately capturing and localizing the changes in the image. 


\begin{figure*}
    \centering
    \includegraphics[width=\linewidth]{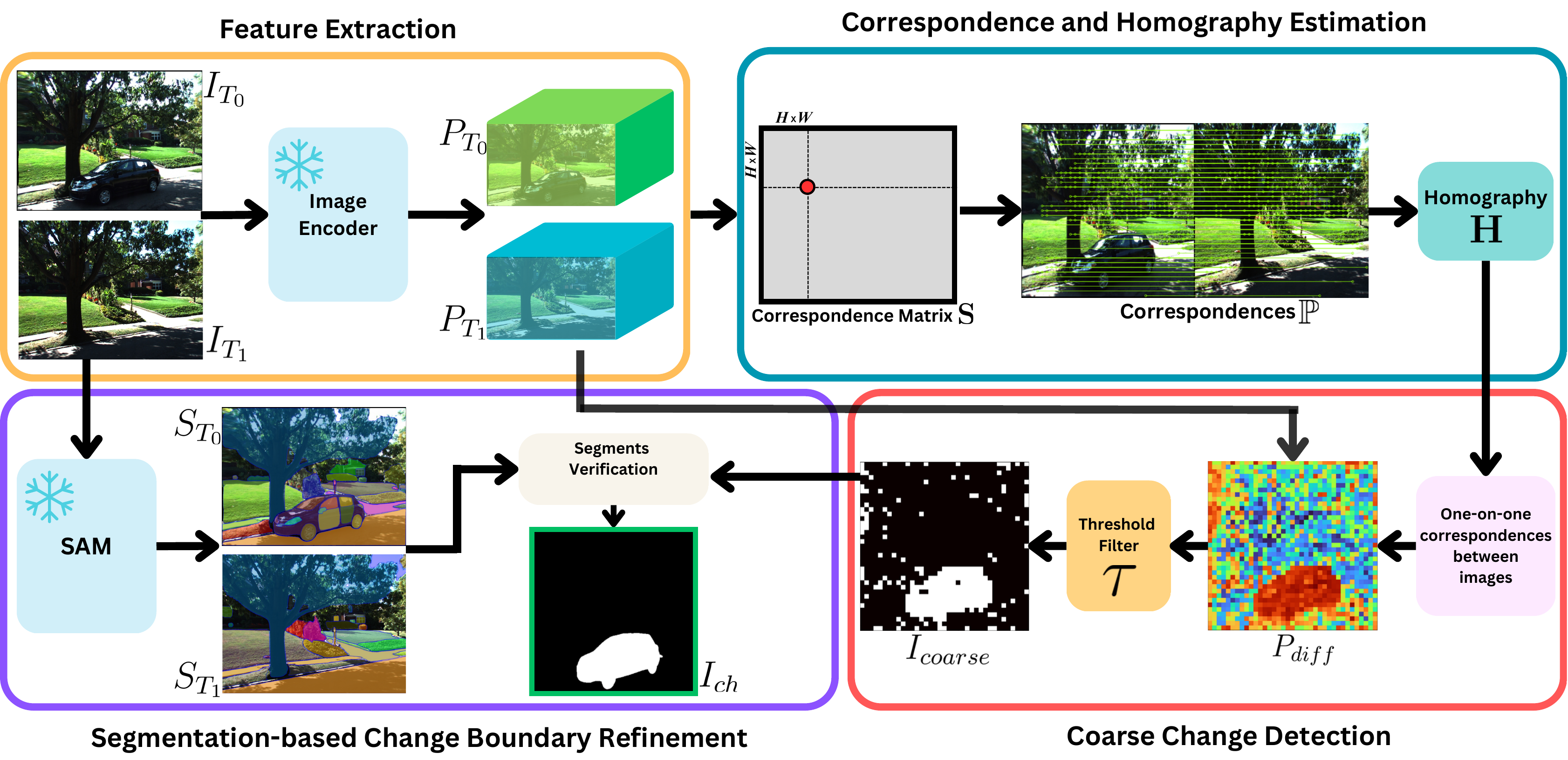}
    \vspace{-6mm}
    \caption{\textbf{Architecture of the ZeroSCD framework}. In ZeroSCD the input images are passed through the image encoder and the patch embeddings are extracted. The homography between the two images is then computed based on the correspondences between the images. Based on the relation between the two images estimated using the homography, a coarse difference map is computed. This difference map identifies patches where changes have occurred. This difference map is then compared with the segmented output of SAM and the segments estimated by SAM that align with the coarse difference map are estimated. The summation of all the segments corresponding to changed regions yields the final change binary mask.} 
    \vspace{-3mm}
    \label{fig:zerocd_archi}
\end{figure*}

\section{Methodology}
\label{sec:methodology}
ZeroSCD is a training-free, zero-shot change detection framework that uses a feature extraction model and a class-agnostic segmentation model to estimate changes between images. The overview of ZeroSCD is shown in Fig.~  \ref{fig:zerocd_archi}, with the framework comprising four main components: Feature Extraction, Correspondence and Homography Estimation, Coarse Change Detection, and Segmentation-based Change Boundary Refinement. First, extracted features are used to estimate correspondences between the images, from which the homography is calculated. Using this, accurate correspondences are determined, and the feature differences in corresponding regions yield a coarse estimate of the changes. Finally, the segments from the segmentation model are compared with the coarse estimate of changes and the individual segments are classified as changed or not changed. These components are detailed in the following subsections.

\subsection{Feature Extraction}
Given the two images $I_{T_{0}}$ and $I_{T_{1}}$ $\in \mathbb{R}^{h \times w \times c}$ captured at times $T_0$ and $T_1$, where $h, w, c$ represent the height, width, and number of channels, respectively, we pass both images through a feature encoder. For feature extraction, we leverage our previous VPR model, PlaceFormer\footnote{For details on PlaceFormer, see “PlaceFormer: Transformer-based Visual Place Recognition using Multi-Scale Patch Selection and Fusion” \cite{kannan2024placeformer}.} \cite{kannan2024placeformer}, which is built on a vision transformer architecture \cite{dosovitskiy2020image} and extracts patch tokens as features. PlaceFormer, originally trained on the diverse Mapillary Street Level Sequences (MSLS) dataset \cite{warburg2020mapillary}, is designed to capture robust features across varying terrains and style conditions. The wide-ranging nature of MSLS ensures that the features extracted by PlaceFormer remain invariant to style variations, enabling ZeroSCD to accurately detect changes even in challenging conditions.

The PlaceFormer encoder produces patch embeddings, $P_{T_{0}}$ and $P_{T_{1}}$ $\in \mathbb{R}^{H \times W \times d}$, which represent the patch-level features for the images 
$I_{T_{0}}$ and $I_{T_{1}}$, respectively. Here, $H$ and $W$ denote the height and width of the patch grid, while $d$ corresponds to the descriptor length of each patch token.

\subsection{Correspondence and Homography Estimation}
The patch embeddings output by the encoder are used to estimate the correspondences between the patches of images $I_{T_{0}}$ and $I_{T_{1}}$ based on their similarity. The correspondence matrix $\mathbf{S}$ $\in \mathbb{R}^{HW \times HW}$ is computed as:

\begin{equation}
    \mathbf{S}_{ij} = \frac{{p}_{T_{0}}^{i} \cdot {p}_{T_{1}}^{j}}{\|{p}_{T_{0}}^{i}\| \cdot \|{p}_{T_{1}}^{j}\|}
\end{equation}

\noindent where $\mathbf{S}_{ij}$ denotes the cosine similarity between the $i$-th patch embedding ${p}_{T_{0}}^{i}$ of $P_{T_{0}}$ and the $j$-th patch embedding ${p}_{T_{1}}^{j}$ of $P_{T_{1}}$. The patch correspondences from $P_{T_{0}}$ to $P_{T_{1}}$ are determined by identifying the maximum value in each row of the similarity matrix $\mathbf{S}$, where the index of the maximum value corresponds to the matching patch in the other image. This approach ensures that each patch in $P_{T_{0}}$ is matched with its most similar counterpart in $P_{T_{1}}$. The yields a set of patch correspondences $\mathbb{P} = \{ (i \rightarrow j) \}$, where $j$-th patch in $P_{T_{1}}$ is the most closest patch to the $i$-th patch in $P_{T_{0}}$.

Using this matching patch set, $\mathbb{P}$, the homography matrix $\mathbf{H}$ is estimated via RANSAC. Each patch is treated as a $2$D point, with its coordinates centered within the patch. For robust homography fitting, an inlier tolerance of $1.25$ times the patch size is applied to account for minor misalignments, ensuring accurate transformation between the two images.

\subsection{Coarse Change Detection}

With the homography matrix $\mathbf{H}$ between the two images now established, the correspondences for all patches from one image to the other can be computed. Let $(u,v)$ represent the $x$ and $y$ coordinates of a patch in image $I_{T_{0}}$. Its corresponding patch $(u^\prime,v^\prime)$ in image 
$I_{T_{1}}$ can be calculated as follows:

With the homography matrix $\mathbf{H}$ between the two images now established, the correspondences for all patches from one image to the other can be determined. Let $(u,v)$ represent the $x$ and $y$ coordinates of a patch in image $I_{T_{0}}$. The corresponding patch coordinates $(u^\prime, v^\prime)$ in image $I_{T_{1}}$ can be calculated as:
\begin{equation}
    \begin{bmatrix}
        u^\prime\\
        v^\prime\\
        1\\
    \end{bmatrix}  = \mathbf{H} \begin{bmatrix}
        u\\
        v\\
        1\\
    \end{bmatrix}
\end{equation}

Let $p_{T_{0}}$ be a patch in image $I_{T_{0}}$ and $p_{T_{1}}$ its corresponding patch in image $I_{T_{1}}$. The Euclidean distance between the descriptors of these two patches is computed, and this process is repeated across all corresponding patches between the images. The resulting heatmap, denoted as ${P_{diff}}$, highlights regions where significant differences have occurred, indicating potential changes over time. To refine this, the patches in ${P_{diff}}$ are compared against a predefined change detection threshold, $\tau$. If the computed distance exceeds $\tau$, the patch is marked as \textit{changed}; otherwise, it is classified as \textit{unchanged}. Repeating this process for all patches in the images identifies the set of changed patches in $I_{T_{1}}$, yielding a coarse estimate of change regions, which forms the coarse change map, $I_{coarse}$.

\subsection{Segmentation-based Change Boundary Refinement}
The images $I_{T_{0}}$ and $I_{T_{1}}$ are processed using the SAM to generate segmented versions, denoted as $S_{T_{0}}$ and $S_{T_{1}}$, respectively. Let ${s_{T_{0}}} \in S_{T_{0}}$ and ${s_{T_{1}}} \in S_{T_{1}}$ represent the sets of individual segment boundaries from these segmented images. Each segment in both the sets is validated to determine whether it belongs to a changed region by comparing it with the coarse change map, $I_{coarse}$, which highlights areas of potential change. The degree of overlap between a segment and the coarsely detected change regions helps in identifying changed segments.

Let $s$ be a segment from either image, and $s_{o}$ its overlap with $I_{coarse}$, computed as $s_{o} = s \cap I_{coarse}$. The ratio of overlap, $\gamma$, is calculated as $\gamma = \frac{area(s_{o})}{area(s)}$. If $\gamma$ exceeds a predefined threshold $\alpha$, the segment is flagged as potentially changed, denoted as $s_{flag}$. To further verify this, the corresponding segment $s_{flag}^{\prime}$ in the other image is located using the homography matrix $\mathbf{H}$. The ratio of overlap between $s_{flag}$ and $s_{flag}^{\prime}$, computed as $\frac{area(s_{flag} \cap s_{flag}^{\prime})}{area(s_{flag})}$, is then checked. If this ratio is below a threshold, indicating significant change, $s_{flag}$ is included in the final binary change mask, $I_{ch}$. This double verification ensures that changes are confirmed both in terms of feature differences and geometrical discrepancies.

This process is applied to all segments from $S_{T_{0}}$ and $S_{T_{1}}$, ultimately producing the final change mask, $I_{ch}$. Segments from $S_{T_{0}}$ that are marked as changed indicate regions that have disappeared or been altered by time $T_{1}$, such as demolished structures or removed objects. Conversely, changes associated with segments from $S_{T_{1}}$ reflect new appearances, indicating additions or newly constructed elements. This distinction allows for a more detailed understanding of the environment's transformation, offering insights into both disappearances and new appearances over time.


\section{Experiments}
\label{sec:expt}

\subsection{Implementation}
\label{sec:implementation}
In the implementation of ZeroSCD, PlaceFormer \cite{kannan2024placeformer} serves as the backbone for feature extraction, chosen for its ability to generate robust and distinctive feature embeddings essential for accurately estimating correspondences between images. Built on a lightweight, small version of the Vision Transformer, PlaceFormer is both efficient and effective for this task. For segmentation, SAM \cite{kirillov2023segment} is employed to produce high-quality segmentation masks. To ensure the detection of all changes, including finer details, SAM was fine-tuned, particularly by adjusting the \textit{points per side} parameter to achieve the appropriate level of segmentation granularity. For change detection, the threshold $\tau$ used to select changed patches was set at $0.65$, and the minimum change ratio threshold $\alpha$ was set at $0.8$ (80\%), ensuring a high degree of precision in identifying alterations. All the implementations and testing were performed on a Nvidia RTX 3090 GPU.

\subsection{Datasets}
\label{sec:dataset}
ZeroCSD was evaluated and benchmarked against other state-of-the-art change detection methods on VL-CMU-CD \cite{alcantarilla2018street} and PCD2025 datasets \cite{jst2015change}. The two datasets were chosen since they cover diverse environments and different level of changes in the images. 

\noindent{\textbf{VL-CMU-CD Dataset} \cite{alcantarilla2018street}} is a change detection dataset derived from the VL-CMU dataset \cite{badino2012real}, which was originally developed for localization tasks. This dataset captures long-term changes, including both structural alterations, such as building demolitions, and style variations, like changes in weather, seasons, and lighting. The dataset consists of $152$ sequences, encompassing a total of $1,362$ image pairs. For training, $97$ sequences with $933$ image pairs are provided, while the testing set includes $54$ sequences with $429$ image pairs. Following standard practices in the field, the images from VL-CMU-CD were resized to a resolution of $512 \times 512$ for evaluation purposes.

\noindent{\textbf{PCD2015 Dataset} \cite{jst2015change}} consists of two distinct subsets: Tsunami and GSV, each presenting unique challenges for change detection. The Tsunami subset features $200$ image pairs captured in the aftermath of a tsunami, highlighting dramatic, large-scale changes in street-level environments. The GSV subset, sourced from Google Street View, includes $92$ image pairs, showcasing more subtle and varied changes typical of urban settings. Since our method is zero-shot and does not require any training data, we evaluate our framework on the entire dataset, unlike other methods that perform fivefold cross-validation. 

\begin{table*}
\centering
\caption{Comparison of ZeroSCD on benchmark dataset against various state-of-the-art methods with the F1-Scores. The best results are highlighted in \textbf{bold}.}
\vspace{-2mm}
\label{tab:cd_results}
\resizebox{0.785\textwidth}{!}{%
\begin{tabular}{lccccccc}
\hline
\multirow{2}{*}{Methods} & \multicolumn{7}{c}{F1-Scores}                                              \\ \cline{2-8} 
                         & VL-CMU-CD \cite{alcantarilla2018street}     &  & Tsunami (PCD2015 \cite{jst2015change}) &  & GSV (PCD2015 \cite{jst2015change}) &  & Average       \\ \hline
CNN-Feat \cite{jst2015change}                 & 40.3          &  & 72.4              &  & 63.9          &  & 58.8          \\
CDNet \cite{alcantarilla2018street}           & 58.2          &  & 77.4              &  & 61.4          &  & 65.6          \\
CosimNet \cite{guo2018learning}               & 70.6          &  & 80.6              &  & 69.2          &  & 73.4          \\
SimUNet \cite{zhang2022resnest}               & 71.4          &  & 82.9              &  & 68.1          &  & 74.1          \\
DOF-CDNet \cite{sakurada2017dense}            & 68.8          &  & 83.8              &  & 70.3          &  & 74.3          \\
DASNet \cite{chen2020dasnet}                  & 72.2          &  & 84.1              &  & 74.5          &  & 76.9          \\
CSCDNet  \cite{sakurada2020weakly}            & 71.0          &  & 85.9              &  & 73.8          &  & 76.9          \\
HPCFNet \cite{lei2020hierarchical}            & 75.2          &  & 86.8              &  & 77.6          &  & 79.86         \\
SimSac \cite{park2022dual}                    & \textbf{75.6} &  & 86.5              &  & 78.2          &  & 80.1          \\ \hline
ZeroSCD (Ours)             & 75.4          &  & \textbf{90.6}     &  & \textbf{82.1} &  & \textbf{82.7} \\ \hline
\end{tabular}%
}
\vspace{-6mm}
\end{table*}

\subsection{Metrics}
\label{sec:metrics}
The accuracy of change detection is evaluated using the F1-score, $F1$ \cite{yang2022scene}. To calculate the F1-score, both precision, $P$, and recall, $R$ must be determined first. These metrics are defined as follows:

\begin{equation}
    F1 = \frac{2 \times P \times R}{P + R}
\end{equation}

where $P= TP/(TP + FP)$ and $R= TP/(TP + FN)$; $TP$ denotes the number of true positives, $FP$ the number of false positives, and $FN$ the number of false negatives. Precision, $P$ represents the proportion of correctly identified changes out of all detected changes, reflecting the algorithm's ability to avoid false detections. Recall, $R$ measures the proportion of actual changes that were correctly identified, indicating the algorithm’s sensitivity to detect changes that truly occurred. The F1-score is the harmonic mean of precision and recall, offering a balanced measure that ranges between 0 and 1, where higher values indicate better performance. Typically, a higher precision suggests a lower rate of false positives, while a higher recall suggests a lower rate of false negatives. The F1-score effectively combines these two metrics, with a higher F1-score signifying superior performance in accurately detecting changes while minimizing both missed detections and false alarms.

\subsection{Comparison with State-of-the-arts}
\label{sec:comp}
ZeroSCD is evaluated against nine other state-of-the-art change detection methods: CNN-Feat \cite{jst2015change}, CDNet \cite{alcantarilla2018street}, CosimNet \cite{guo2018learning}, SimUNet \cite{zhang2022resnest}, DOF-CDNet \cite{sakurada2017dense}, DASNet \cite{chen2020dasnet}, CSCDNet \cite{sakurada2020weakly}, HPCFNet \cite{lei2020hierarchical}, and SimSac \cite{park2022dual} on benchmark datasets. All of these methods leverage CNNs in various forms for feature extraction, typically utilizing well-known architectures like VGG-16 \cite{simonyan2014very}, ResNet-18 \cite{he2016deep}, and UNet \cite{huang2020unet }. The extracted features from these networks are subsequently processed through additional layers to compute the change masks, which highlight the differences between image pairs.

\section{Results}
\label{sec:results}

\subsection{Quantitative Results}
In the PCD2015 dataset, ZeroSCD demonstrates superior performance compared to all other methods across both the Tsunami and GSV subsets. In the Tsunami subset, ZeroSCD surpasses the second-best performing method, SimSac, by a substantial margin of $4.1\%$, while in the GSV subset, it outperforms by $3.9\%$. Both subsets feature structural changes within urban landscapes, highlighting ZeroSCD's effectiveness in detecting such changes robustly. Notably, SimSac was specifically trained on these respective datasets, whereas ZeroSCD achieves its performance in a zero-shot manner, underscoring its ability to generalize without the need for task-specific training.

In the VL-CMU-CD dataset, ZeroSCD performs comparably to SimSac, the best-performing method in this benchmark. The VL-CMU-CD dataset is particularly challenging as it includes not only structural changes but also variations in illumination, weather, and seasons. ZeroSCD's high performance indicates its capability to accurately identify structural changes while remaining resilient to environmental factors such as lighting and seasonal variations. This resilience is critical for real-world applications where environmental conditions can vary widely. Overall, ZeroSCD outperforms the second-best method, SimSac, by an average margin of $2.6\%$, demonstrating not only its superior ability to detect structural changes but also its adaptability across different conditions without the need for dataset-specific tuning. This positions ZeroSCD as a highly versatile and effective change detection model, capable of delivering reliable results across diverse environments and conditions, making it well-suited for practical applications in dynamic urban settings.

\subsection{Ablation Study}
We perform multiple ablation experiments to further affirm design choices made in ZeroSCD. 

\noindent{\textbf{Change Detection Threshold, $\tau$}.}
The patch embeddings corresponding to regions with potential changes are identified using the change detection threshold, $\tau$. This threshold helps ensure that only significant changes are detected, minimizing false positives caused by minor feature mismatches. Table \ref{tab:thresh} presents ablation results evaluating the impact of various threshold values on the VL-CMU-CD dataset. Our experiments show that increasing $\tau$ improves the F1-score by detecting more patches with substantial changes. The F1-score peaks at $\tau = 0.65$, after which it begins to decline as further increases in the threshold start to miss valid changes. Hence, we select $\tau = 0.65$ as the default for all experiments. 

\begin{table}[h]
\centering
\caption{Ablation study on various thresholds for change detection on VL-CMU-CD dataset.}
\label{tab:thresh}
\resizebox{0.45\textwidth}{!}{%
\begin{tabular}{|c|l|l|l|l|c|}
\hline
Threshold, $\tau$ & 0.5 & 0.55 & 0.6 & \textbf{0.65} & 0.7 \\ \hline
F1-Score  & 70.9 & 72.4 & 74.5 & \textbf{75.4} & 74.9     \\ \hline
\end{tabular}%
}
\end{table}

\noindent{\textbf{Different Backbones.}}
Vision foundational models like DINOv2 \cite{oquab2023dinov2} are highly effective at addressing a wide range of vision challenges, even in their pre-trained state. In this ablation study, we explored using different variants of DINOv2 as the backbone for feature extraction, replacing PlaceFormer. As shown in Table \ref{tab:bb_cd_abl}, DINOv2 achieved performance comparable to PlaceFormer, highlighting the scalability of our zero-shot pipeline across different backbones. However, DINOv2 slightly underperformed, likely due to its lack of fine-tuning on street-view images, a domain where PlaceFormer excels due to its targeted training.

\begin{table}[!h]
\centering
\caption{Ablation study on various backbones for feature extraction on VL-CMU-CD dataset.}
\label{tab:bb_cd_abl}
\resizebox{0.285\textwidth}{!}{%
\begin{tabular}{cc}
\hline
Backbone        & F1-Score \\ \hline
PlaceFormer     & 75.4     \\
DINOv2 ViT-S/14 & 70.8     \\
DINOv2 ViT-B/14 & 74.0     \\ \hline
\end{tabular}%
}
\vspace{-4mm}
\end{table}

\subsection{Qualitative Results}
\begin{figure}[t]
    \centering
    \includegraphics[width=\linewidth]{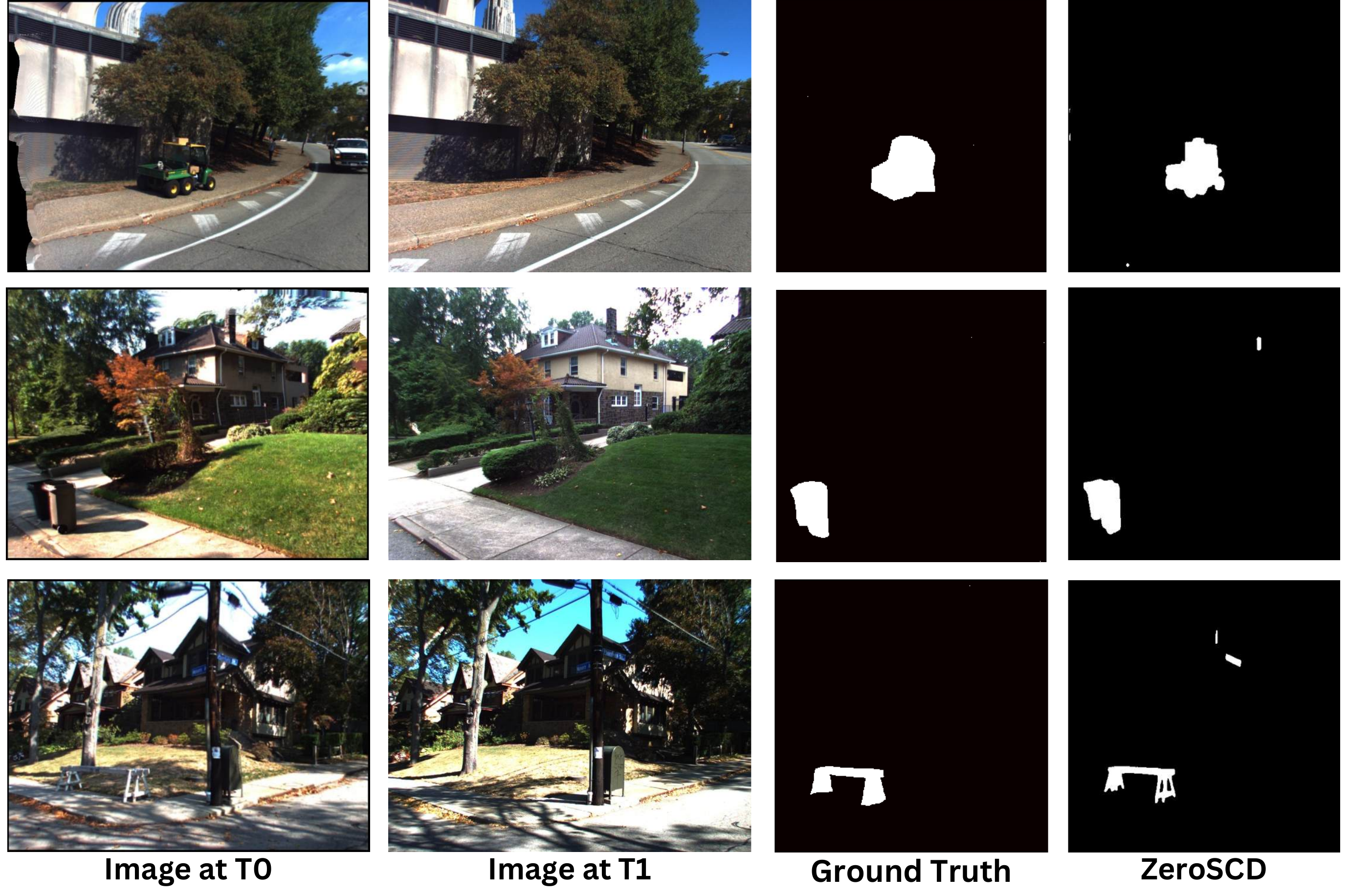}
    \vspace{-6mm}
    \caption{Binary change masks generated by our method on various VL-CMU-CD dataset along with the input images and the ground truth.}  
    \label{fig:quali_1}
\end{figure}

In Fig.~ \ref{fig:quali_1}, we present the binary masks generated by ZeroSCD on images from the VL-CMU-CD dataset. The first row shows a truck that was removed over time, and ZeroSCD accurately captures the truck's boundaries. Similarly, in the second row, a removed trash can is detected, although some noise is also present. While our method is generally robust to illumination changes, certain lighting variations occasionally introduce noise. In the third row, a bench that has been removed is detected with high precision, even capturing the gaps between its legs, which are not reflected in the ground truth. From the first and third rows, it is evident that ZeroSCD provides more precise boundary detection than the rough outlines in the ground truth masks. This improved accuracy is attributed to the use of SAM for generating detailed boundaries. Moving forward, we aim to further explore ZeroSCD's ability to generate accurate boundaries, potentially utilizing it to produce refined ground truth annotations for other tasks.

\begin{figure}[t]
    \centering
    \includegraphics[width=\linewidth]{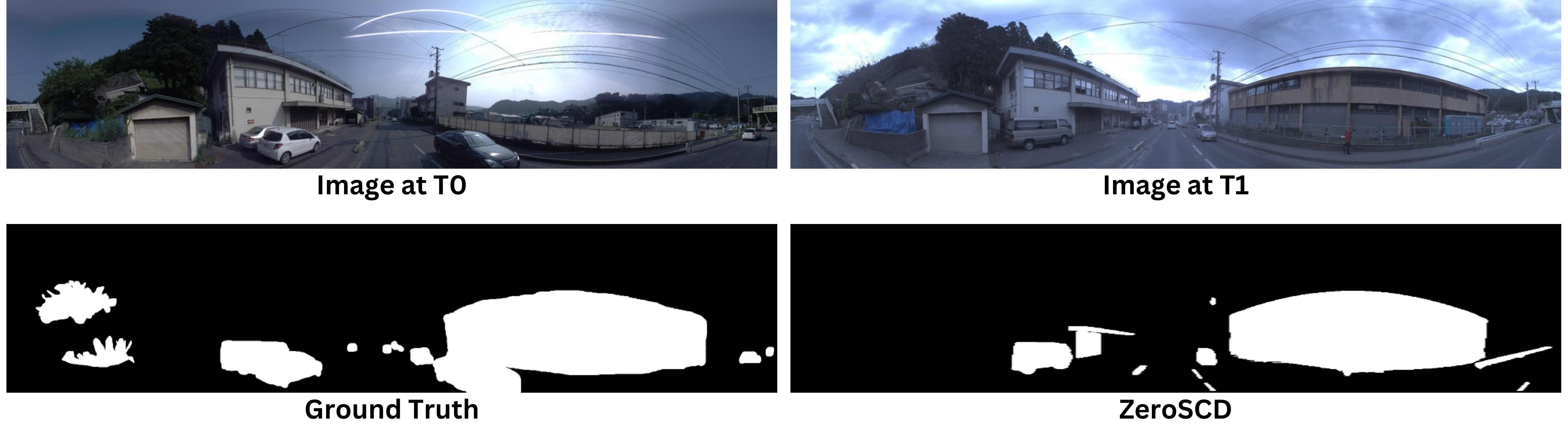}
    \vspace{-6mm}
    \caption{Binary change mask generated by our method for an image pair from the Tsunami dataset along with the input images and the ground truth.}  
    \vspace{-3mm}
    \label{fig:quali_2}
\end{figure}

Fig.~ \ref{fig:quali_2} illustrates the results of our method on an image pair from the Tsunami dataset. While our approach successfully detects changes in the buildings and a nearby car, it overlooks alterations in the vegetation and distant vehicles. Changes in vegetation were often missed because the features extracted for trees, both with and without leaves, appeared too similar. To address this limitation, we plan to improve the robustness of our framework by further refining the VPR model to better distinguish such subtle changes. Additionally, the vision transformer's patch resolution caused smaller objects, such as cars in the distance, to go undetected. To improve detection in these cases, we aim to integrate specialized small object detection techniques into the framework.

\section{Conclusion}
\label{sec:conclusion}
In this paper, we introduced ZeroSCD, a novel zero-shot, training-free approach for scene change detection. By leveraging pre-trained models for place recognition and semantic segmentation, ZeroSCD extracts robust features and segmentations to detect and localize changes between two images of the same scene—without requiring additional training or annotated data. Despite its zero-shot nature, ZeroSCD achieves state-of-the-art performance on multiple change detection benchmarks, offering a scalable and efficient solution for real-world applications, particularly in autonomous vehicle map updates.

However, ZeroSCD’s reliance on two separate models for feature extraction and segmentation introduces computational overhead, making it slower than other methods. In future work, we aim to explore unified foundational models that handle both tasks to reduce this load. Additionally, we plan to expand ZeroSCD's adaptability to new domains, such as aerial imagery, to broaden its applicability to diverse change detection scenarios.

\newpage
\bibliographystyle{IEEEtran}
\bibliography{references}
\end{document}